\documentclass[11pt,a4paper]{article}
\usepackage{acl2015}
\usepackage{times}
\usepackage{url}
\usepackage{latexsym}

\usepackage{graphicx}
\usepackage{algorithm}
\usepackage[noend]{algpseudocode}
\usepackage{multirow}

\usepackage{amsmath}
\usepackage{amssymb}
\usepackage{bm}
\usepackage{balance}

\usepackage{longtable}
\usepackage{rotating}

\usepackage{CJKutf8}

\title{LCSTS: A Large Scale Chinese Short Text Summarization Dataset}

\author{  Baotian Hu~~~~~ Qingcai Chen~~~~~ Fangze Zhu
\\
\\
\begin{tabular}{c}
 { Intelligent Computing Research Center}    \\
 {Harbin Institute of Technology, Shenzhen Graduate School}            \\
 {\tt \{baotianchina,qingcai.chen, zhufangze123\}@gmail.com} \\
\end{tabular}
}

\begin{document}
\maketitle
\begin{abstract}
Automatic text summarization is widely regarded as the highly difficult problem, partially because of the lack of large text summarization data set. Due to the great challenge of constructing the large scale summaries for full text, in this paper, we introduce a large corpus of Chinese short text summarization dataset constructed from the Chinese microblogging website Sina Weibo, which is released to the public\footnote{http://icrc.hitsz.edu.cn/Article/show/139.html}. This corpus consists of over 2 million real Chinese short texts with short summaries given by the author of each text. We also manually tagged the relevance of 10,666 short summaries with their corresponding short texts. Based on the corpus, we introduce recurrent neural network for the summary generation and achieve promising results, which not only shows the usefulness of the proposed corpus for short text summarization research, but also provides a baseline for further research on this topic.

\end{abstract}
\section{Introduction}
\vspace{-2pt}
Nowadays, individuals or organizations can easily share or post information to the public on the social network. Take the popular Chinese microblogging website (Sina Weibo) as an example, the People's Daily, one of the media in China, posts more than tens of weibos (analogous to tweets) each day. Most of these weibos are well-written and highly informative because of the text length limitation (less than140 Chinese characters). Such data is regarded as naturally annotated web resources~\cite{sun}. If we can mine these high-quality data from these naturally annotated web resources, it will be beneficial to the research that has been hampered by the lack of data.

\begin{figure}[!tb]
\centering
\includegraphics[width=\columnwidth]{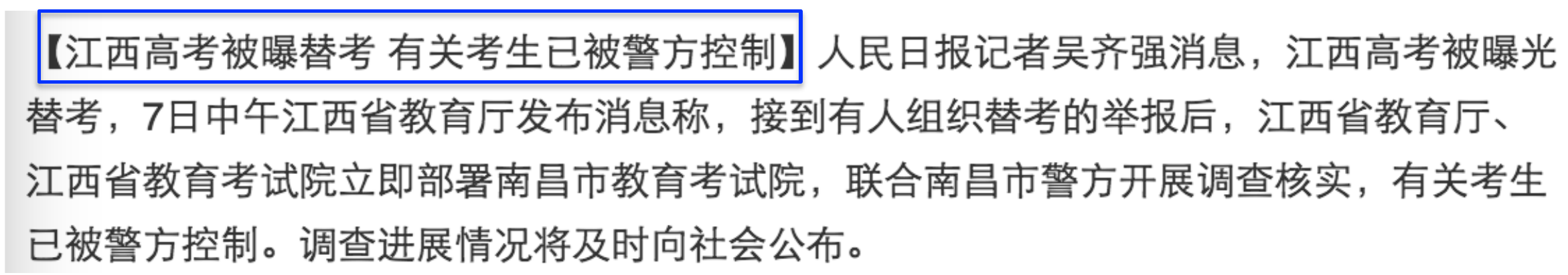}
\vspace{-20pt}
\caption{A Weibo Posted by People's Daily. }
\label{weibo-example}
\vspace{-15pt}
\end{figure}

\begin{figure*}[!tb]
\centering
\includegraphics[width=1.9\columnwidth]{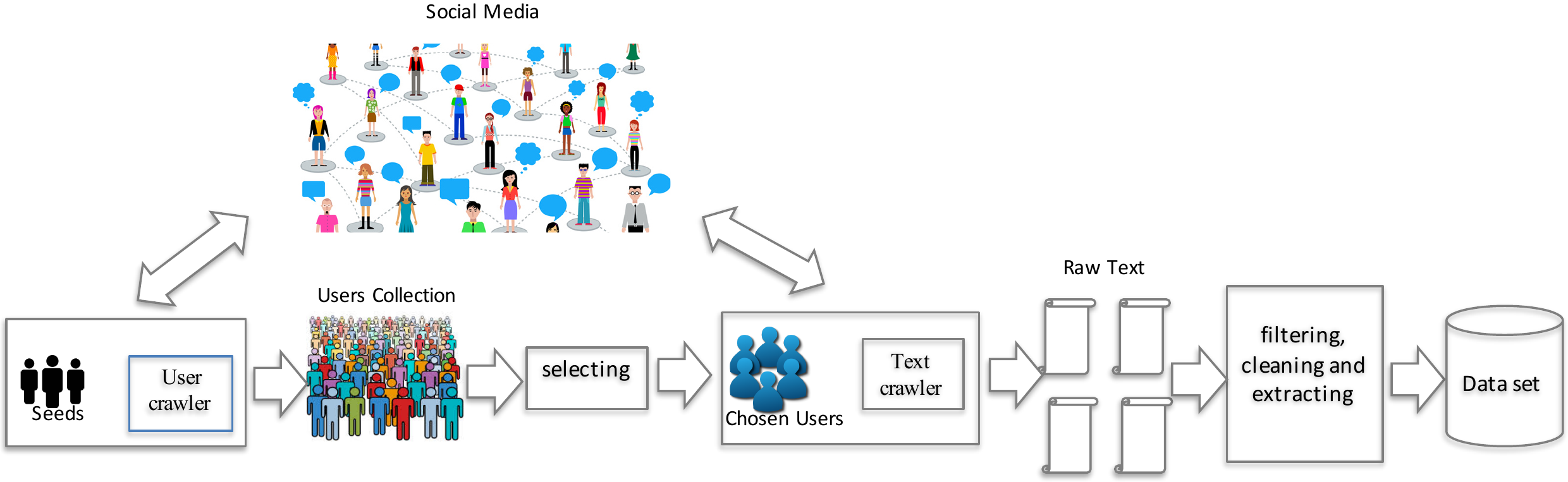}
\vspace{-10pt}
\caption{Diagram of the process for creating the dataset.}
\label{abc}
\vspace{-10pt}
\end{figure*}

In the Natural Language Processing (NLP) community, automatic text summarization is a hot and difficult task. A good summarization system should understand the whole text and re-organize the information to generate coherent, informative, and significantly short summaries which convey important information of the original text~\cite{hovy1998},~\cite{survey_2007}. Most of traditional abstractive summarization methods  divide the process into two phrases~\cite{lidong}. First, key textual elements are extracted from the original text by using unsupervised methods or linguistic knowledge. And then, unclear extracted components are rewritten or paraphrased to produce a concise summary of the original text by using linguistic rules or language generation techniques. Although extensive researches have been done, the linguistic quality of abstractive summary is still far from satisfactory. Recently, deep learning methods have shown potential abilities to learn representation~\cite{my_nips2014,my_acl2015} and generate language~\cite{groundhog,google} from large scale data by utilizing GPUs. Many researchers realize that we are closer to generate abstractive summarizations by using the deep learning methods. However, the publicly available and high-quality large scale summarization data set is still very rare and not easy to be constructed manually. For example, the popular document summarization dataset DUC\footnote{http://duc.nist.gov/data.html}, TAC\footnote{http://www.nist.gov/tac/2015/KBP/} and TREC\footnote{http://trec.nist.gov/} have only hundreds of human written English text summarizations. The problem is even worse for Chinese. In this paper, we take one step back and focus on constructing \textbf{LCSTS}, the \textbf{L}arge-scale \textbf{C}hinese \textbf{S}hort \textbf{T}ext \textbf{S}ummarization dataset by utilizing the naturally annotated web resources on Sina Weibo. Figure~\ref{weibo-example} shows one weibo posted by the People's Daily. In order to convey the import information to the public quickly, it also writes a very informative and short summary (in the blue circle) of the news. Our goal is to mine a large scale, high-quality short text summarization dataset from these texts.

This paper makes the following contributions: (1) We introduce a large scale Chinese short text summarization dataset. To our knowledge, it is the largest one to date; (2) We provide standard splits for the dataset into large scale training set and human labeled test set which will be easier for benchmarking the related methods; (3) We explore the properties of the dataset and sample 10,666 instances for manually checking and scoring the quality of the dataset; (4) We perform recurrent neural network based encoder-decoder method on the dataset to generate summary and get promising results, which can be used as one baseline of the task.

\section{Related Work}
\vspace{-2pt}

Our work is related to recent works on automatic text summarization and natural language processing based on naturally annotated web resources, which are briefly introduced as follows.

\textbf{Automatic Text Summarization} in some form has been studied since 1950. Since then, most researches are related to extractive summarizations by analyzing the organization of the words in the document~\cite{survey_2011}~\cite{ibmluhn}; Since it needs labeled data sets for supervised machine learning methods and labeling dataset is very intensive, some researches focused on the unsupervised methods~\cite{unspervised_1}. The scale of existing data sets are usually very small (most of them are less than 1000). For example, DUC2002 dataset contains 567 documents and each document is provided with two 100-words human summaries. Our work is also related to the headline generation, which is a task to generate one sentence of the text it entitles. Colmenares et.al construct a 1.3 million financial news headline dataset written in English for headline generation~\cite{headline}. However, the data set is not publicly available.

 \textbf{Naturally Annotated Web Resources based Natural Language Processing} is proposed by Sun~\cite{sun}. Naturally Annotated Web Resources is the data generated by users for communicative purposes such as web pages, blogs and microblogs. We can mine knowledge or useful data from these raw data by using marks generated by users unintentionally. Jure et.al track 1.6 million mainstream media sites and blogs and mine a set of novel and persistent temporal patterns in the news cycle~\cite{sun1}. Sepandar et.al use the users' naturally annotated pattern `we feel' and `i feel' to extract the `Feeling' sentence collection which is used to collect the world's emotions. In this work, we use the naturally annotated resources to construct the large scale Chinese short text summarization data to facilitate the research on text summarization.

\section{Data Collection}
\vspace{-2pt}

A lot of popular Chinese media and organizations have created accounts on the Sina Weibo. They use their accounts to post news and information. These accounts are verified on the Weibo and labeled by a blue `V'. In order to guarantee the quality of the crawled text, we only crawl the verified organizations' weibos which are more likely to be clean, formal and informative.  There are a lot of human intervention required in each step. The process of the data collection is shown as Figure~\ref{abc} and summarized as follows:

1) We first collect 50 very popular organization users as seeds. They come from the domains of politic, economic, military, movies, game and etc, such as People's Daily, the Economic Observe press, the Ministry of National Defense and etc. 2) We then crawl fusers followed by these seed users and filter them by using human written rules such as  the user must be blue verified,  the number of followers is more than 1 million and etc. 3) We use the chosen users and text crawler to crawl their weibos. 4) we filter, clean and extract (short text, summary) pairs.  About 100 rules are used to extract high quality pairs. These rules are concluded by 5 peoples via carefully investigating of the raw text.  We also remove those paris, whose short text length is too short (less than 80 characters) and length of summaries is out of [10,30].

\section{Data Properties}
\vspace{-2pt}

\begin{figure}[!tb]
\centering
\includegraphics[width=\columnwidth]{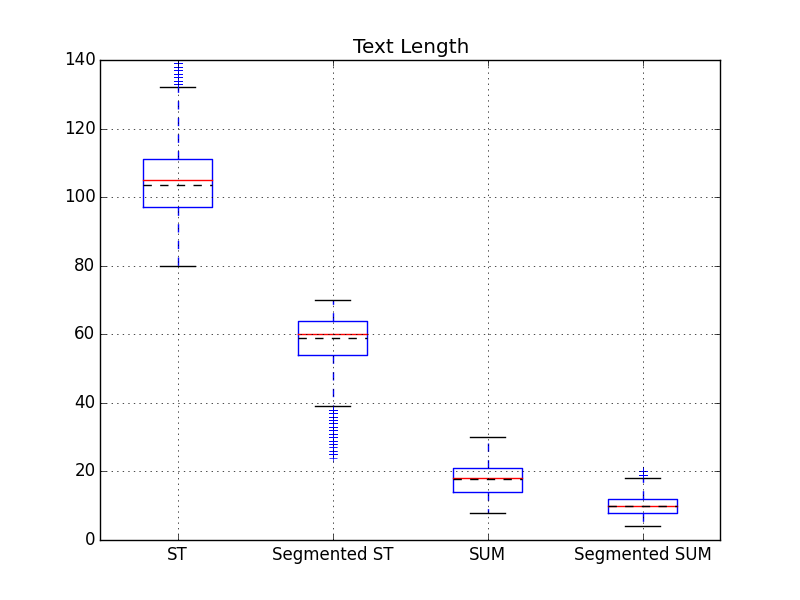}
\vspace{-25pt}
\caption{Box plot of lengths for short text(ST), segmented short text(Segmented ST), summary(SUM) and segmented summary(Segmented SUM). The red line denotes the median, and the edges of the box the quartiles. }
\label{tongji}
\vspace{-15pt}
\end{figure}

The dataset consists of three parts shown as Table~\ref{tabel_sta} and the length distributions of texts are shown as Figure~\ref{tongji}.

\textbf{Part I} is the main content of LCSTS that contains 2,400,591 (short text, summary) pairs. These pairs can be used to train supervised learning model for summary generation.

\begin{figure}[!tb]
\centering
\includegraphics[width=\columnwidth]{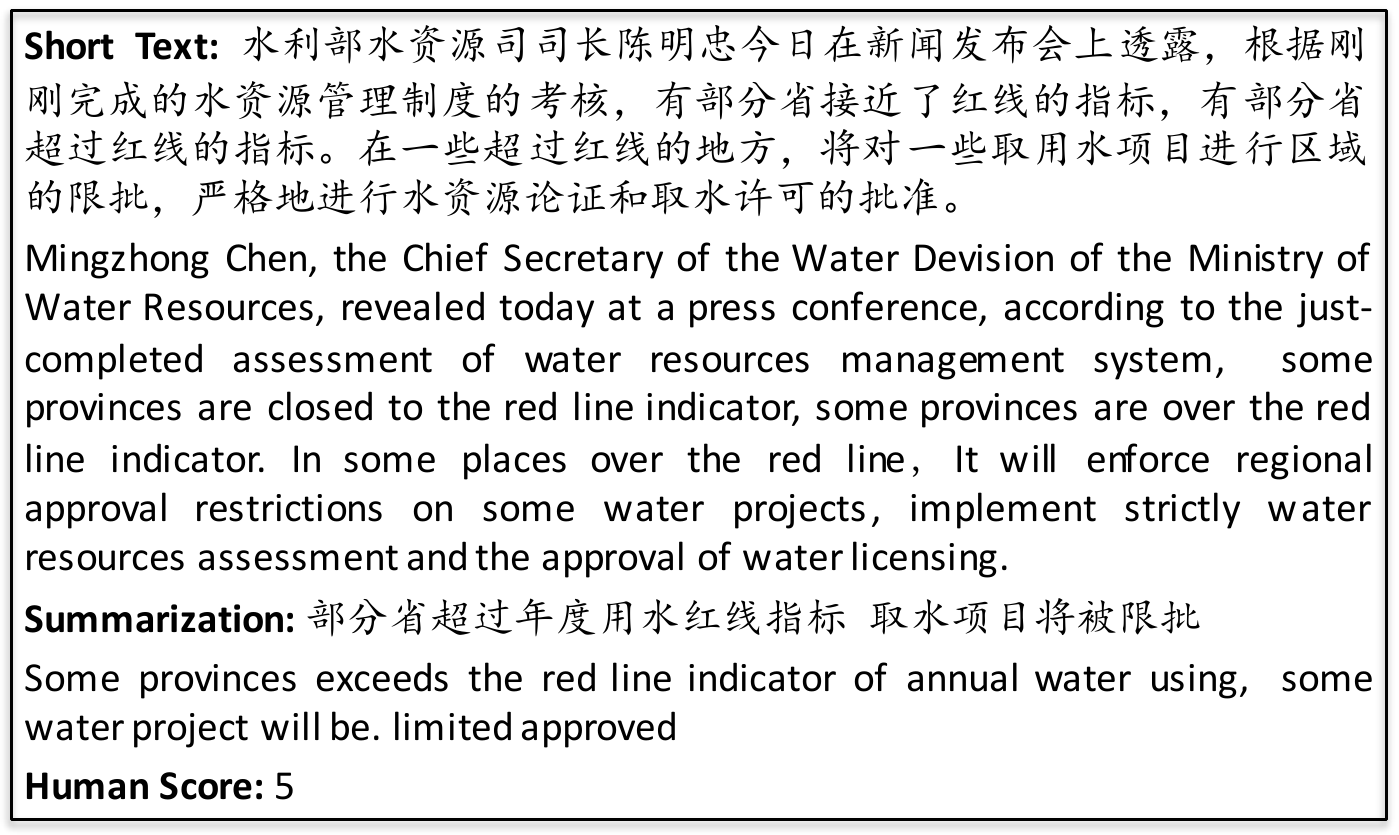}

\includegraphics[width=\columnwidth]{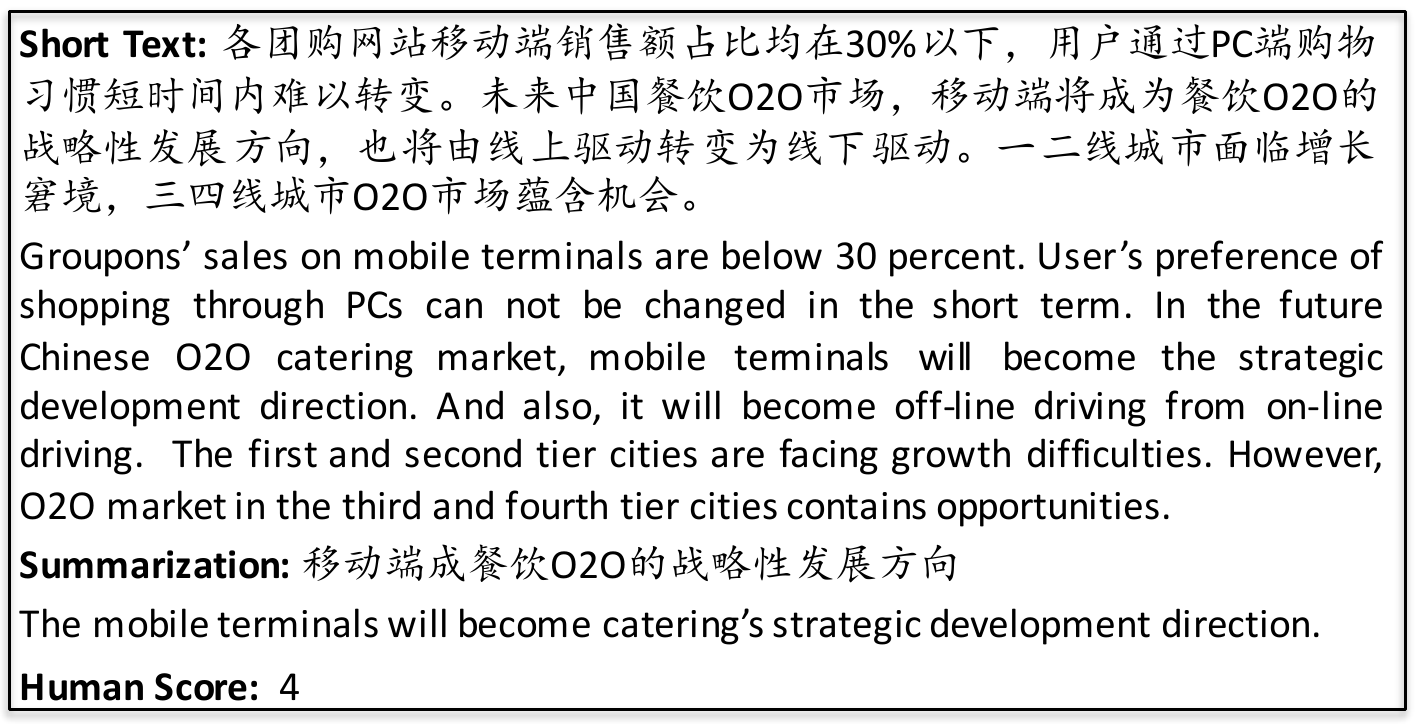}

\includegraphics[width=\columnwidth]{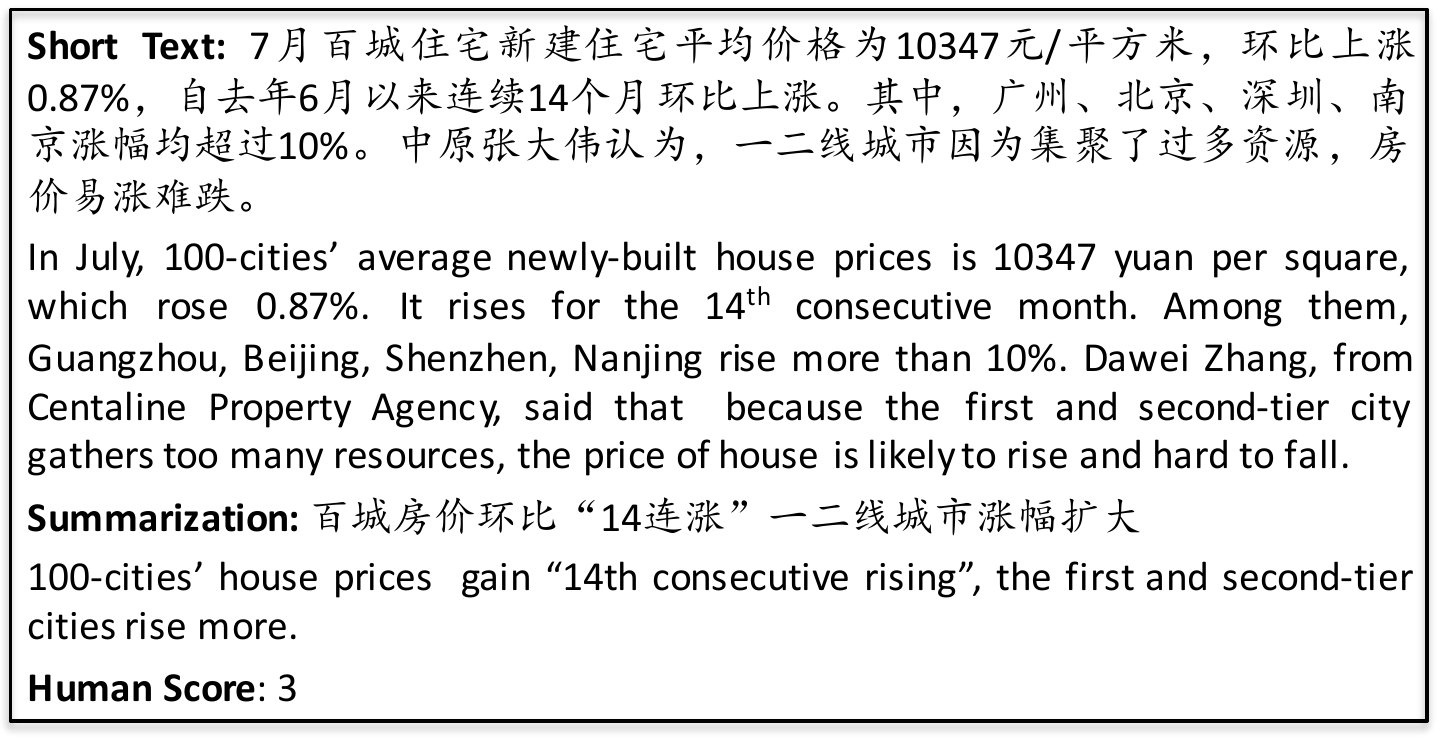}

\includegraphics[width=\columnwidth]{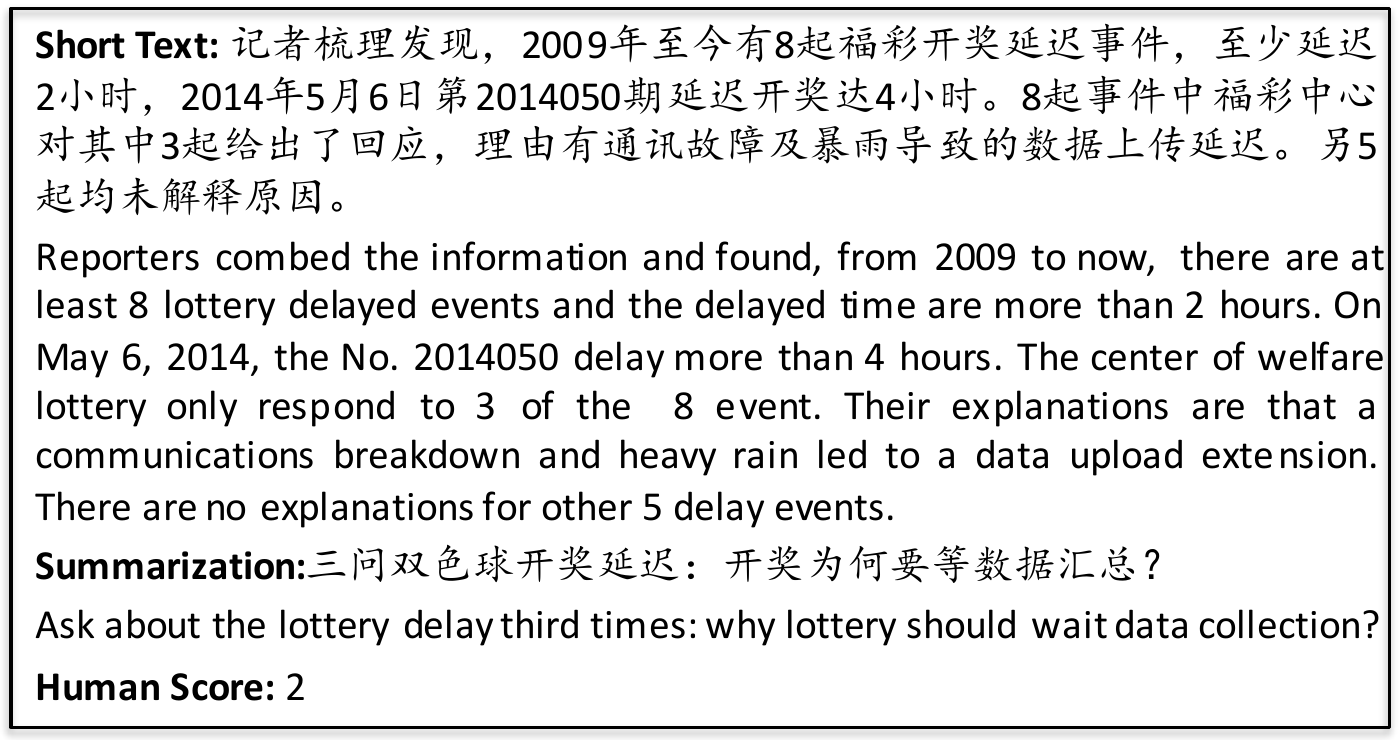}

\includegraphics[width=\columnwidth ]{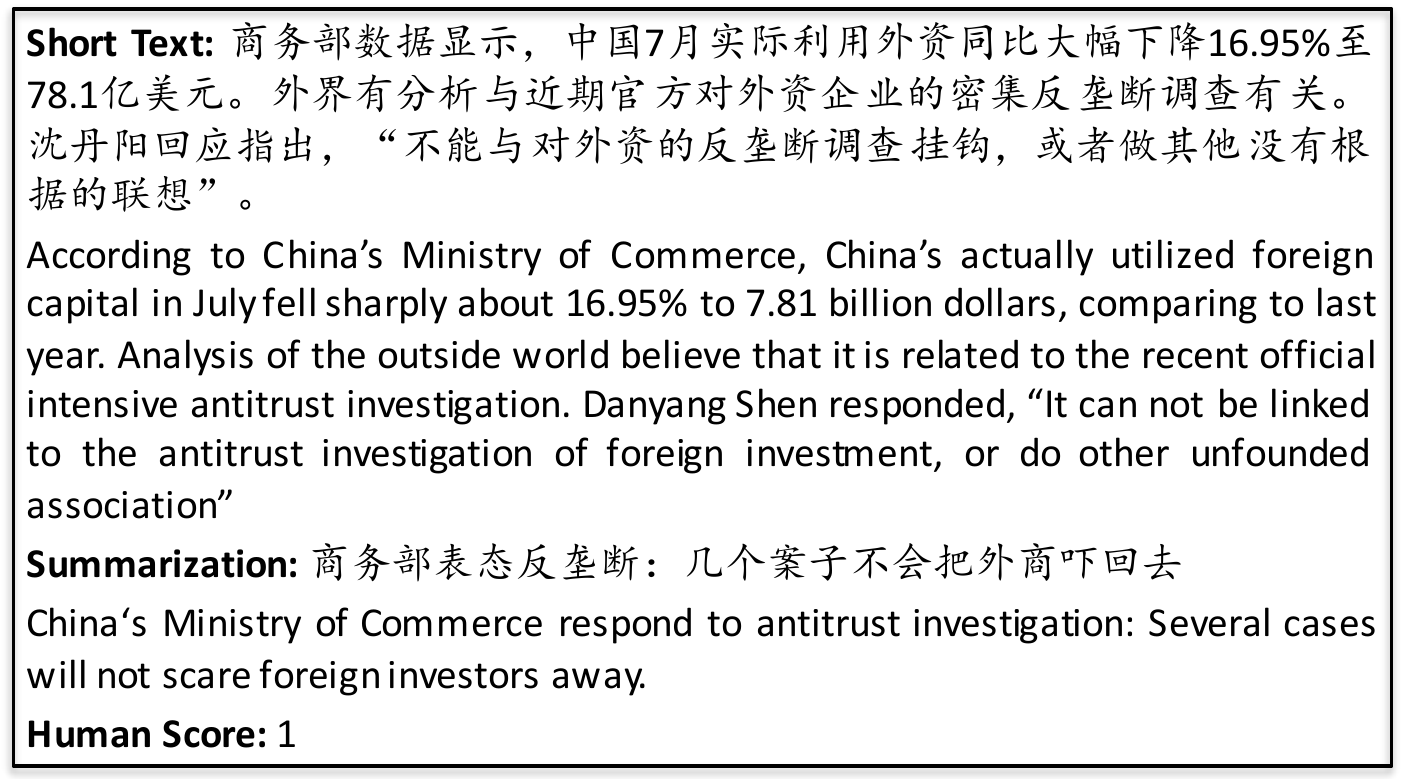}
\vspace{-25pt}
\caption{Five examples of different scores.}
\label{fig_scores}
\vspace{-20pt}
\end{figure}

\textbf{Part II} contains the 10,666 human labeled (short text, summary) pairs with the score ranges from 1 to 5 that indicates the relevance between the short text and the corresponding summary. `1' denotes `` the least relevant '' and `5' denotes ``the most relevant''.  For annotating this part,  we recruit 5 volunteers, each pair is only labeled by one annotator. These pairs are randomly sampled from Part I and are used to analysize the distribution of pairs in the Part I. Figure~\ref{fig_scores} illustrates examples of different scores. From the examples we can see that pairs scored by 3, 4 or 5 are very relevant to the corresponding summaries. These summaries are highly informative, concise and significantly short compared to original text. We can also see that many words in the summary do not appear in the original text, which indicates the significant difference  of our dataset from sentence compression datasets. The summaries of pairs scored by 1 or 2 are highly abstractive and relatively hard to conclude the summaries from the short text. They are more likely to be headlines or comments instead of summaries. The statistics show that the percent of score 1 and 2 is less than 20\% of the data, which can be filtered by using trained classifier.

\textbf{Part III} contains 1,106 pairs.
 For this part,  3 annotators label the same 2000 texts and we extract the text with common scores. This part is independent from Part I and Part II. In this work, we use pairs scored by 3, 4 and 5 of this part as the test set for short text summary generation task.

\begin{table}[!h]
\begin{center}
\begin{tabular}{|c|c|c|}
\hline
Part I & \multicolumn{2}{c|}{2,400,591} \\
\hline
\multirow{6}{*}{Part II} & Number of Pairs & 10,666\\
\cline{2-3}
& Human Score 1 & 942 \\
 \cline{2-3}
  & Human Score 2 & 1,039  \\
 \cline{2-3}
  & Human Score 3 & 2,019  \\
   \cline{2-3}
  & Human Score 4 & 3,128  \\
   \cline{2-3}
  &Human Score  5 & 3,538  \\
\hline

\multirow{6}{*}{Part III} & Number of Pairs &1,106 \\
 \cline{2-3}
& Human Score 1 & 165 \\
 \cline{2-3}
  & Human Score 2 & 216  \\
 \cline{2-3}
  & Human Score 3 & 227  \\
   \cline{2-3}
  & Human Score 4 & 301  \\
   \cline{2-3}
  &Human Score  5 & 197  \\
\hline
\end{tabular}
\end{center}
\vspace{-10pt}
\caption{Data Statistics}
\label{tabel_sta}
\vspace{-5pt}
\end{table}

\section{Experiment}
\vspace{-2pt}

Recently, recurrent neural network (RNN) have shown powerful abilities on speech recognition~\cite{speech}, machine translation~\cite{google} and automatic dialog response~\cite{lifeng}. However, there is rare research on the automatic text summarization by using deep models. In this section, we use RNN as encoder and decoder to generate the summary of short text. We use the Part I as the training set and the subset of Part III, which is scored by 3, 4 and 5, as test set.
\begin{figure}[!tb]
\centering
\includegraphics[width=\columnwidth]{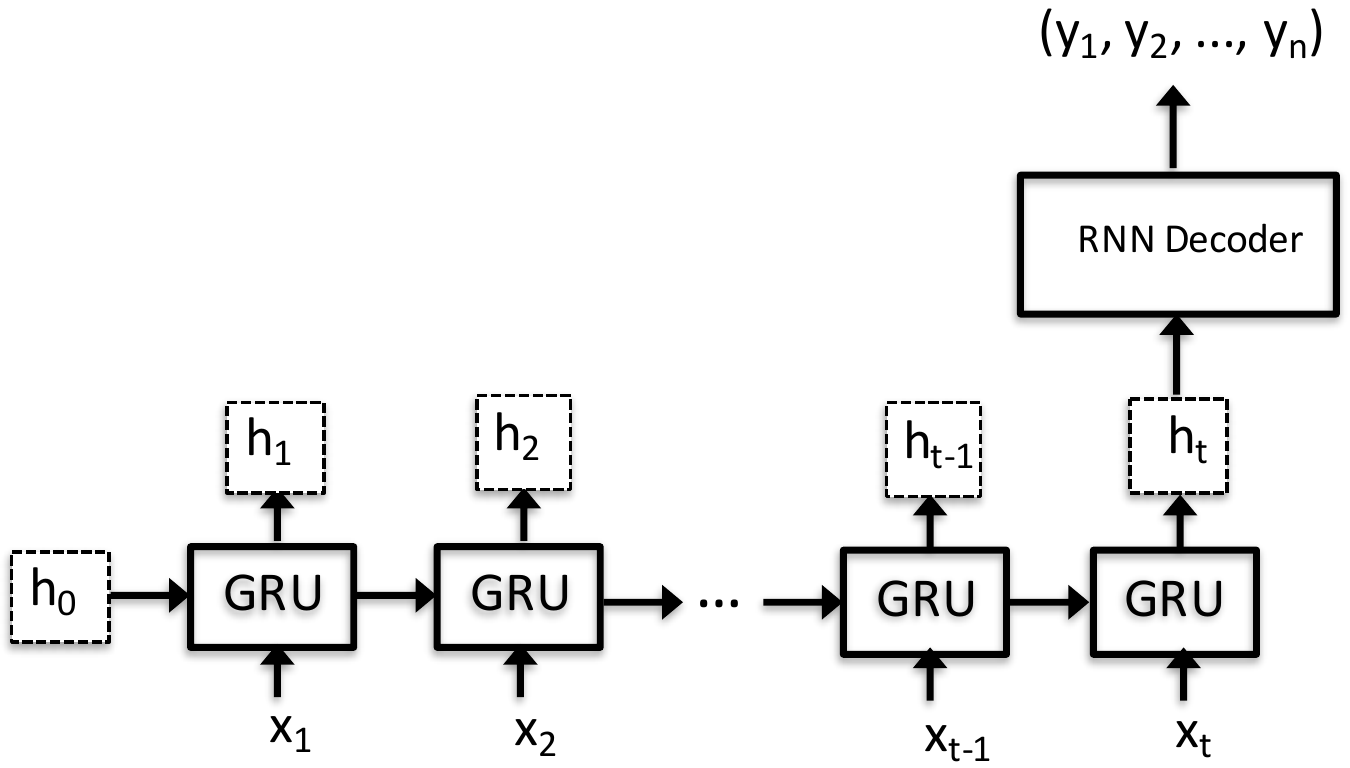}
\vspace{-25pt}
\caption{The graphical depiction of  RNN encoder and decoder framework without context. }
\label{model_no_context}
\vspace{-1pt}
\end{figure}

\begin{figure}[!tb]
\centering
\includegraphics[width=\columnwidth]{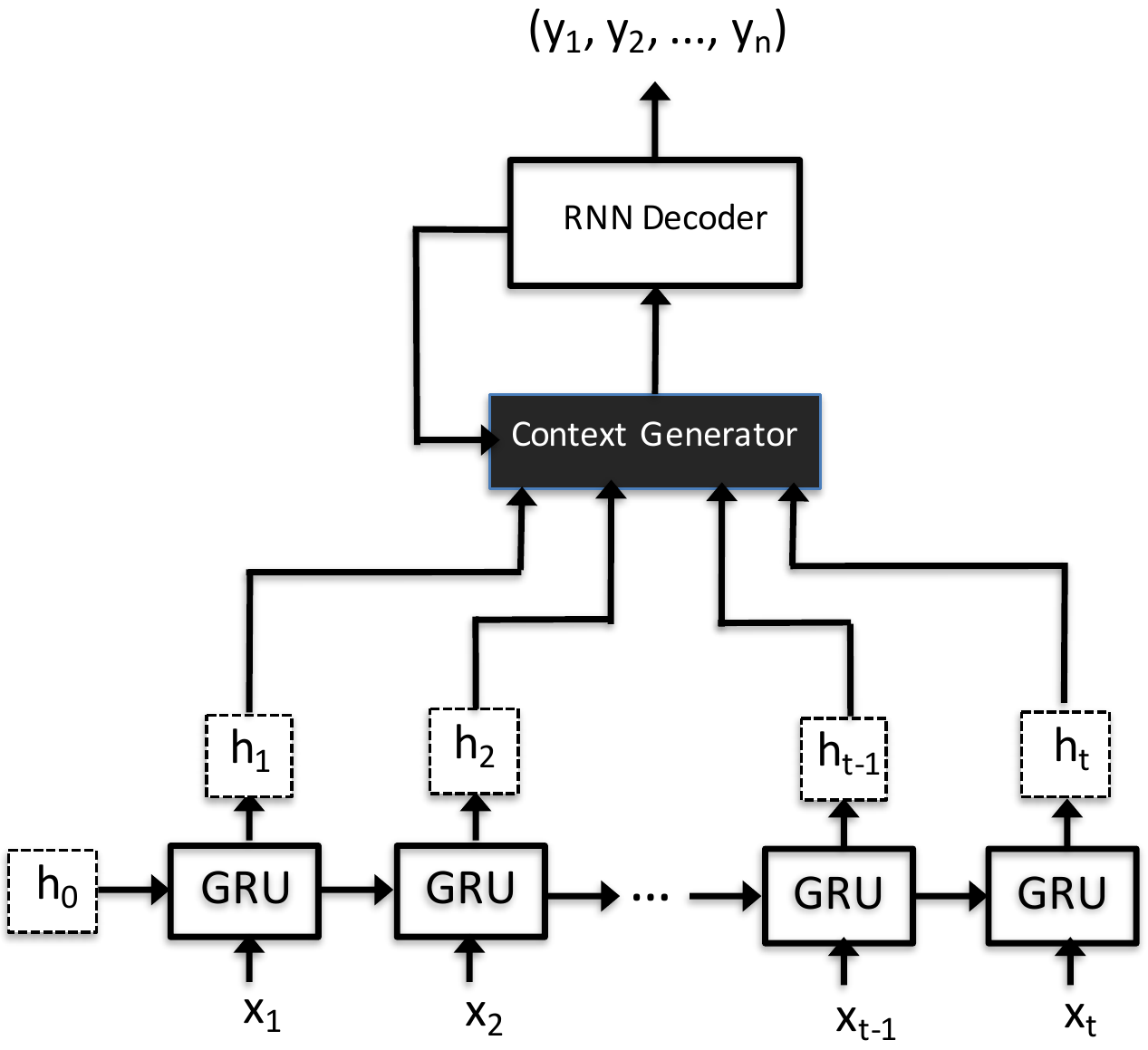}
\vspace{-25pt}
\caption{The graphical depiction of the RNN encoder and decoder framework with context. }
\label{model_context}
\vspace{-5pt}
\end{figure}

Two approaches are used to preprocess the data: 1) character-based method, we take the Chinese character as input, which will reduce the vocabulary size to 4,000. 2) word-based method, the text is segmented into Chinese words by using jieba\footnote{https://pypi.python.org/pypi/jieba/}. The vocabulary is limited to 50,000. We adopt two deep architectures, 1) The local context is not used during decoding. We use the RNN  as encoder and it's last hidden state as the input of decoder, as shown in Figure~\ref{model_no_context}; 2) The context is used during decoding, following~\cite{groundhog}, we use the combination of all the hidden states of encoder as input of the decoder, as shown in Figure~\ref{model_context}.  For the RNN, we adopt the gated recurrent unit (GRU) which is proposed by ~\cite{gatedRNN} and has been proved comparable to LSTM~\cite{lstm_vs_gru}. All the parameters (including the embeddings) of the two architectures are randomly initialized and ADADELTA~\cite{adadelta} is used to update the learning rate.  After the model is trained, the beam search is used to generate the best summaries in the process of decoding and the size of beam is set to 10 in our experiment.

\begin{table}[!h]
\begin{center}
\begin{tabular}{|c|c|c|c|c|}
\hline
model & data& R-1 & R-2 & R-L \\
\hline
\multirow{2}{*}{RNN} & Word& 0.177 &0.085 &0.158 \\
 \cline{2-5}
  & Char& 0.215 & 0.089 & 0.186  \\
\hline

 \multirow{2}{*}{RNN context} & Word& 0.268 & 0.161 &0.241\\
 \cline{2-5}
  & Char& \textbf{0.299} &\textbf{0.174} & \textbf{0.272}  \\
  \hline

 \end{tabular}
\end{center}
\caption{The experiment result: ``Word'' and  ``Char'' denote the word-based and  character-based input respectively.}
\label{eval}
\vspace{-10pt}
\end{table}

\begin{figure}[!tb]
\centering
\includegraphics[width=\columnwidth]{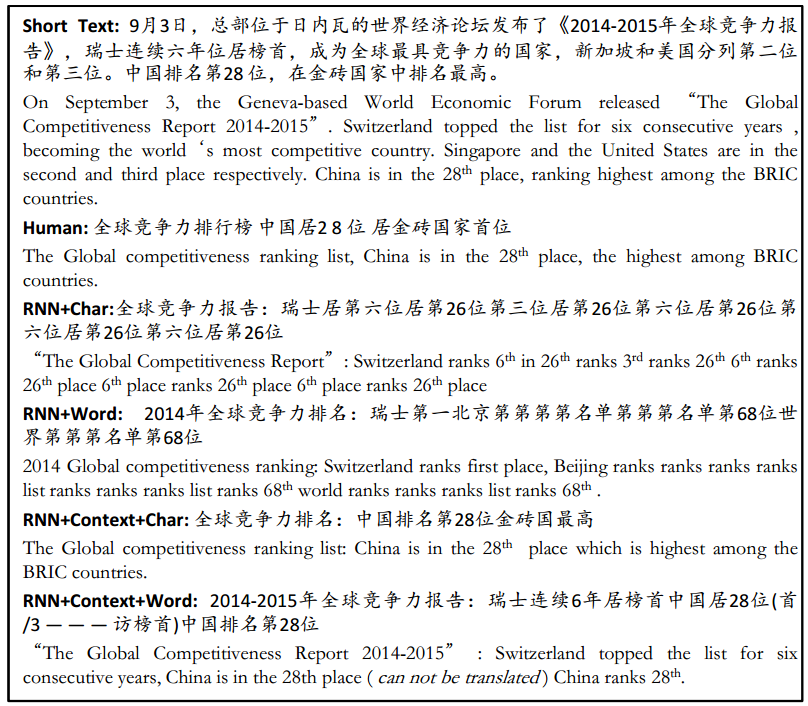}
\vspace{-25pt}
\caption{An example of the generated summaries.}
\label{fig_exp}
\vspace{-12pt}
\end{figure}

For evaluation, we adopt the ROUGE metrics proposed by~\cite{rouge_2003}, which has been proved strongly correlated with human evaluations. ROUGE-1, ROUGE-2 and ROUGE-L are used. Because the standard Rouge package~\footnote{http://www.berouge.com/Pages/default.aspx} is used for evaluating English summarization systems, we transform the Chinese words to numerical IDs to adapt to the systems. All the models are trained on the GPUs tesla M2090 for about one week.Table~\ref{eval} lists the experiment results. As we can see in Figure~\ref{fig_exp}, the summaries generated by RNN with context are very close to human written summaries, which indicates that if we feed enough data to the RNN encoder and decoder, it may generate summary almost from scratch.

\begin{figure}[!tb]
\centering
\includegraphics[width=\columnwidth]{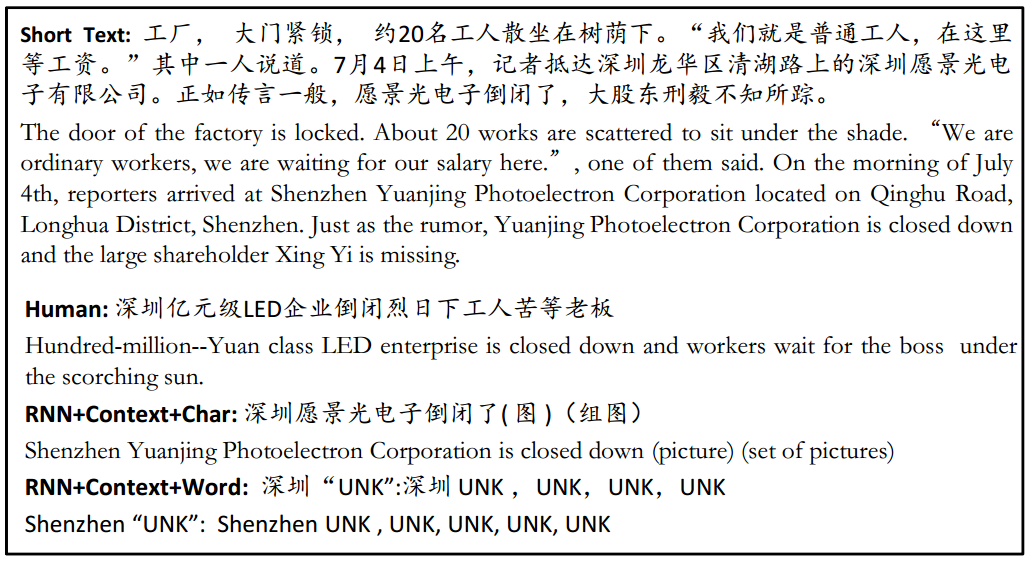}
\vspace{-25Pt}
\caption{An example of the generated summaries with UNKs.}
\label{fig_exp_unks}
\vspace{-12pt}
\end{figure}

The results also show that the RNN with context outperforms RNN without context on both character and word based input. This result indicates that the internal hidden states of the RNN encoder can be combined to represent the context of words in summary. And also the performances of the character-based input outperform the word-based input.  As shown in Figure~\ref{fig_exp_unks},  the summary generated by RNN with context by inputing the character-based short text is relatively good, while the  the summary generated by RNN with context on word-based input contains many UNKs. This may attribute to that the segmentation may lead to many UNKs in the vocabulary and text such as the person name and organization name. For example,  \begin{CJK}{UTF8}{gkai} ``愿景光电子'' \end{CJK} is a company name which is not in the vocabulary of word-based RNN, the RNN summarizer has to use the UNKs to replace the  \begin{CJK}{UTF8}{gkai} ``愿景光电子'' \end{CJK} in the process of decoding.

\vspace{-2pt}
\section{Conclusion and Future Work}
\vspace{-2pt}
We constructed a large-scale Chinese short text summarization dataset and performed RNN-based methods on it, which achieved some promising results. This is just a start of deep models on this task and there is much room for improvement. We take the whole short text as one sequence, this may not be very reasonable, because most of short texts contain several sentences. A hierarchical RNN~\cite{jiwei} is one possible direction. The rare word problem is also very important for the generation of the summaries, especially when the input is word-based instead of character-based.
It is also a hot topic in the neural generative models  such as neural translation machine(NTM)~\cite{rareword}, which can benefit to this task. We also plan to construct a large document summarization data set by using naturally annotated web resources.

\vspace{-2pt}
\section*{Acknowledgments}
 \vspace{-5pt}
This work is supported by National Natural Science Foundation of China: 61473101, 61173075 and 61272383, Strategic Emerging Industry Development Special Funds of Shenzhen: JCYJ20140417172417105, JCYJ20140508161040764 and JCYJ20140627163809422. We thank to Baolin Peng, Lin Ma, Li Yu and the anonymous reviewers for their insightful comments.

%\clearpage
\vspace{-10pt}

\bibliographystyle{acl}
\balance
\bibliography{my}

\end{document}